\documentclass[conference]{IEEEtran}
\IEEEoverridecommandlockouts
\usepackage{cite}
\usepackage{amsmath,amssymb,amsfonts}
\usepackage{algorithmic}
\usepackage{graphicx}
\usepackage{textcomp}
\usepackage{xcolor}
\def\BibTeX{{\rm B\kern-.05em{\sc i\kern-.025em b}\kern-.08em
		T\kern-.1667em\lower.7ex\hbox{E}\kern-.125emX}}
\usepackage[normalem]{ulem}
\usepackage{listings}

\usepackage{xspace}
\newcommand{\tueba}{T\"{u}Ba-D/Z\xspace}

\begin{document}
	
	\title{Resource-Size matters: Improving Neural Named Entity Recognition with Optimized Large Corpora\\ 
	}
	
	\author{\IEEEauthorblockN{Sajawel Ahmed, Alexander Mehler}
		\IEEEauthorblockA{\textit{Text Technology Lab} \\
			\textit{Goethe University Frankfurt}\\
			Frankfurt, Germany\\
			\{sahmed, mehler\}@em.uni-frankfurt.de}
	}
	
	\maketitle
	
	\begin{abstract} 
		This study improves the performance of neural named entity recognition by a margin of up to 11\% in F-score on the example of a low-resource language like German, thereby outperforming existing baselines and establishing a new state-of-the-art on each single open-source dataset. Rather than designing deeper and wider hybrid neural architectures, we gather all available resources and perform a detailed  optimization and grammar-dependent morphological processing consisting of lemmatization and part-of-speech tagging prior to exposing the raw data to any training process. We test our approach in a threefold monolingual experimental setup of a) single, b) joint, and c) optimized training and shed light on the dependency of downstream-tasks on the size of corpora used to compute word embeddings.
	\end{abstract}
	
	
	\section{Introduction}
	\textit{Named Entity Recognition} (NER) is a crucial part of various \textit{Natural Language Processing} (NLP) tasks like entity linking, relation extraction, machine reading and ultimately \textit{Question Answering} (QA). 
	With the recent rise of neural networks, much emphasis has been put on high-resource languages like English or Chinese leading to fast advancements of many foundational tasks, in particular NER which in many areas reaches near-human performance for these languages \cite{glample2016, ouyang2017chinese}. 
	However, for other, less-resource languages like German, their neural NER counterparts did not attract similar attention from the deep learning community, leading to lower performance by a margin of up to 11\% F-score.
	
	In this paper, we look for the reasons and take steps towards solving them.
	By example of German we bridge the current gap between the performance of neural NER for different languages and bring the performance to a new state-of-the-art. 
	We report evidence that the inferior quality of German text data and its small size are the major reasons for the observed lack of progress.
	
	To tackle this problem, we use a larger corpus for training the foundational word embeddings, namely \textit{Leipzig40} \cite{Goldhahn2012BuildingLM} (including the whole German Wikipedia till 2016) combined with the \textit{WMT 2010 German monolingual training data} \cite{callison2010findings}, and contrast its use with the \textit{COW corpus} \cite{Schaefer2015b}, the largest collection of German texts extracted from web documents with over 617 Mio.\ sentences. 
	Besides, we bring all scattered (open-source) resources of annotated NER datasets for German together which are to date available, prepare and merge them to increase the amount of the final training data. This includes the major NER datasets of \textit{CoNLL-2003} \cite{tjong2003introduction} and \textit{GermEval-2014} \cite{Benikova2014NoStaDNE}, and the smaller datasets of \textit{Europarl-2010} \cite{faruqui10:_training} and of \textit{EuropeanaNewspapers-2016} \cite{neudecker16.110}. To this collection, we add the dataset of T\"ubingen Treebank (\textit{\tueba}) \cite{telljohann2006stylebook}, which to the knowledge of the authors is utilized the first time for the task of neural NER.
	
	It is an increasing scientific practice to make models open source accessible.
	New models appear almost daily, for example in the \textit{Deep Learning} (DL) community. 
	As a consequence, changing existing models and trying out different hybrid setups is getting a scientific practice involving more and more scientists. 
	This is advantageous, since attempts to improve existing models can contribute to their validation. 
	However, it is often forgotten that \textit{data is the gold of scientist}: 
	it is the availability of limited resources that leads to significant improvements in various areas such as CoNLL, SNLI \cite{bowman2015large} and SQuAD \cite{rajpurkar2016squad} for the tasks NER, \textit{natural language inference} and QA and stand behind the recent success of neural networks in NLP.
	Therefore it is important to consider sufficient available resources, to annotate them according to the task and to optimize them if necessary. 
	This task is often time-consuming and costly. 
	The present paper deals with assessing the impact of resources to NER by example of a rather low-resource language like German.
	We show the influence of different training sets on the performance of neural NER, of different combinations of these data sets and above all of different levels of their preprocessing.
	We deal with the aspect of resource optimization with regard to lemmatization and \textit{Part-of-Speech} (POS) tagging and analyze their influence besides the training of word embeddings and task-specific neural networks.
	Our main finding is: an increase of size and quality of the (task-independent) word embedding corpus and of the (task-specific) training dataset leads to a significant improvement of sequence labeling tasks like NER, which can be larger than just an amendment of the underlying neural architecture. For the future of neural NER by example of less- or low-resource languages this means: 
	collecting unlabeled corpora for training morphology-dependent, high quality embeddings is a good alternative to increase the performance of downstream-tasks.

	The remainder of the paper is organized as follows: Section 2 reviews related work, Section 3 presents a sketch of the underlying model, Section 4 describes our threefold experimental setup of a) single, b) joint, and c) resource optimized training, Section 5 reports and discusses our results, and, finally, Section 6 draws a conclusion. 
	\section{Related Work}
	Compared to high-resource languages, comparatively less emphasis has been put on the task of neural NER by example of German. Noteworthy work has been done so far only by \cite{nreimers2014} on GermEval and by \cite{glample2016} on CoNLL; both will be used as baselines here. Reimers et al. \cite{nreimers2014} were among the first to apply neural networks to German NER. However, they did not consider GermEval in combination with CoNLL. Apart from them, the remaining studies (predominantly conducted by non-native speakers) consider this task as a side product of dealing with various other languages. In this way, the state-of-the-art on German neural NER has been established by \cite{glample2016} in 2016. 
	
	Gillick et al. \cite{Gillick2016MultilingualLP} consider German as a variant in a multilingual training setup while additionally considering the datasets of two Germanic languages (English and Dutch) and one Romanic language (Spanish) from the CoNLL shared task; as a result, they reach 76.22 \% F-score. However, for the single training on the German part of CoNLL they stay below \cite{nreimers2014}.
	
	From the point of view of resource optimization, the recent work of \cite{klimek:2018:germeval:analysis} is worth mentioning. 
	Klimek et al.\ also observe the gap between the languages and therefore carry out a detailed analysis of the difficulties for the German NER task using the GermEval data set as an example. 
	They come to the conclusion that \textit{``the task of German NER could benefit from integrating morphological processing''} \cite{klimek:2018:germeval:analysis}. 
	To this end, we start our analysis and apply our designed morphological processing approach to all text corpora and NER datasets.
		
	\section{Model}\label{sec:Model}
	Our neural model consist of two separately trained components: a) foundational word embeddings, modeling the general knowledge from large unlabeled text corpora, and b) task-specific neural networks, modeling the domain knowledge from the labeled training data. In this section, both components are presented briefly.
	\paragraph{Word Embeddings} The language model of continuous space word representations (\textit{word2vec}) \cite{mikolov2013distributed} and its variations by \cite{levy2014dependency,komninos2016dependency} are the foundations of most ongoing research in NLP with neural networks. Based on the context, the model embeds words, phrases or sentences into high dimensional vector spaces. In such a space, the semantics of associations of words and phrases are captured to such an extent that algebraic operations lead to meaningful relationships (e.g.\ $\text{vec(\textit{king})} - \text{vec(\textit{man})} + \text{vec(\textit{woman})} \approx \text{vec(\textit{queen})}$ \cite{mikolov2013distributed}). This property is immensely useful for our  application. 
	We use the model of \textit{word2vec} and its extension \textit{wang2vec} \cite{Ling:2015:naacl} which explores syntactic data and, thus, better suites the task of NER.
	\paragraph{Neural Model}We give a brief sketch of the neural model \textit{LSTM-CRF} which we use throughout this paper. The model is similar to the one used in \cite{glample2016}, which goes back to the works of \cite{Chiu2016NamedER,Huang2015BidirectionalLM,collobert2011natural}. We use a neural model consisting of stacked LSTM and CRF layers. The \textit{base layer} is made of two parts: (i) a preprocessing sublayer generating the character-based embeddings with a cell of forward and backward LSTMs (\textit{biLSTM}) \cite{graves2013speech}, and the word embeddings from the input sentence, (ii) followed by an encoding sublayer again with a cell of a biLSTM extracting features and generating compressed hidden representations. The \textit{prediction layer} is made of CRFs and takes the previous hidden representations to finally produce the \textit{Named Entity} (NE) tag predictions.
	
	Let $ (w_1,\ldots, w_{N_s}) = [w_i]$ be the list of words of a sentence from the input corpus of texts. Furthermore, let $ (c_{i,1}, \ldots, c_{i,N_{w_i}})= [c_{i,l}] $ be the list of characters of the word $ w_i $ consisting of $ N_{w_i} $ characters with $ c_{i,l} $ being its $l$\textsuperscript{th} character. For a given word $ w_i $ and its NE-tag (gold label) $ t_i \in $ \textit{\{PER, LOC, ORG, MISC, O\}} the data flow within the neural network is as follow:
	\begin{IEEEeqnarray}{c}
		\text{char2vec}(c_{i,l})\mapsto \vec{c_{i,l}}\\
		\text{biLSTM}([\vec{c_{i,l}}]) \mapsto \vec{h^c_i}\\
		\text{word2vec}(w_i)\mapsto \vec{w_i}\\
		\text{biLSTM}([(\vec{w_i}, \vec{h^c_i})]) \mapsto [\vec{h^w_i}]\\
		\text{CRF}([\vec{h^w_i}]) \mapsto [t_i]
	\end{IEEEeqnarray}
	where char2vec is a (randomly initialized) lookup table for embedding all characters into a corresponding vector space, and $ (\vec{w_i},\vec{h^c_i}) $  is the concatenation of the embedding vector  of word $w_i$ and its character-based hidden representation. 
	The model is trained to predict the NE-tag $ t_i $ for each word after seeing the whole input sentence at once.
	\section{Experimental Setup}
	
	\subsection{Datasets} 
	In order to evaluate our model of Section \ref{sec:Model} for neural NER on German data, we put emphasis on the major datasets of CoNLL (German part) and GermEval. However, more German resources are available that have so far gone unnoticed in the DL community. In Table \ref{tab-datasets-ner}, we gather all these NER datasets, which are to date freely accessible, and list them along their number of sentences. Additionally, for each dataset the total number of NE tokens is provided along the four categories from the standards defined in the CoNLL shared task 2003 (CoNLL format). Table \ref{tab-datasets-ner} shows that the \tueba dataset is the largest of these, both in terms of the number of sentences and of tokens, ideally fitting to the needs of deep neural networks.	
	\begin{table}[htbp]
		\caption{NER Datasets}
		\begin{center}
			\resizebox{0.5\textwidth}{!}{			\begin{tabular}{|c|c||c|c|c|c|}
					\hline
					\textbf{Corpus} & \textbf{\textit{Sent.}}&\textbf{\textit{PER}}&\textbf{\textit{LOC}}&\textbf{\textit{ORG}}&\textbf{\textit{MISC}} \\
					\hline\hline
					CoNLL-2003&\textcolor{white}{0}18,024&\textcolor{white}{0}\textbf{8,309}&\textcolor{white}{0}7,864&\textcolor{white}{0}7,621&\textcolor{white}{0}4,748\\
					\hline
					Europarl-2010&\textcolor{white}{00}4,395&\textcolor{white}{00}514&\textcolor{white}{00}724&\textcolor{white}{00}874&\textcolor{white}{00}\textbf{966}\\
					\hline
					GermEval-2014&\textcolor{white}{0}31,300&16,204&\textbf{16,675}&12,885&9,254\\
					\hline
					Europ.Newsp.-2016&\textcolor{white}{00}8,879&\textcolor{white}{0}\textbf{7,914}&\textcolor{white}{0}6,143&\textcolor{white}{0}2,784&\textcolor{white}{0000}3\\
					\hline
					\tueba-2018&\textbf{104,787}&\textbf{55,746}&28,582&32,224&12,865\\ 
					\hline
				\end{tabular}
			}
			\label{tab-datasets-ner}
		\end{center}
	\end{table}
	
	\paragraph{Preprocessing of Training Data} 	
	Apart from CoNLL, most copora had to be further processed to fit the CoNLL format. For GermEval, we consider only the top-level NE, refraining from nested NE to stay in line with the remaining datasets. As a tagging scheme, we preferred the BIO (IOB2) scheme, as it has been shown to perform better \cite{Reimers2017ReportingSD}. All datasets are given in the BIO scheme, except CoNLL (IOB1) and Europarl (IOB1), which we converted into the target scheme.
		
	For EuropeanaNewspapers, we take the two datasets written in standard German orthography, namely \textit{enp\_DE.lft.bio} and \textit{enp\_DE.sbb.bio} based on historic newspapers from the Dr.\ Friedrich Tessmann Library and the Berlin State Library, respectively, and omit the Austrian historic newspapers which use a different orthography, differing heavily from the former samples.
	The original dataset is not provided in the 4-column CoNLL format, which writes each word of a sentence horizontally along its lemma, POS tag and  NE-label, and separates each sentence by an empty newline. Therefore, we convert the data into our target format by using \textit{spaCy V2.0}\footnote{http://spacy.io} which by its recent release supports preprocessing German texts by providing language models for sentence boundary detection, lemmatization and POS tagging.
	
	For \tueba, we extracted the NE-tags from the \textit{tuebadz-11.0-conll2010} version. In the case of nested NE, we use a filtering heuristics to extract the longest spanning NE, which allowed us to get more robust training data, not splitting well known entities into parts (e.g.\ \textit{[Goethe Universit\"at Frankfurt]}\_ORG vs. \textit{[Goethe]}\_PER \textit{Universit\"at} \textit{[Frankfurt]}\_LOC). We converted the tagging scheme of \tueba to our target format. Lastly, to allow comparisons with other NER datasets, we mapped the NE category \textit{Geo Political Entity} (GPE) to \textit{LOC}. 
	
	\paragraph{Data Splitting \& Merging} For CoNLL and Germ\-Eval we use the splits as provided in the original datasets. Further, we split \tueba into train/dev/test sets according to the common ratio of 80/10/10 percentages. Due to the smaller size of the Europarl und the EuropeanaNewspapers datasets, we did not consider them for the first experimental setup of single training, rather we merged them with the training data for the second experimental setup of joint training. For this setup, we aligned all datasets by mapping the NE category \textit{OTH} to \textit{MISC} to fit to the CoNLL format. In this way, we generated the currently largest training dataset for German NER of a size of $133,258$ sentences.\footnote{CoNLL (12,152) + GermEval (24,000) + Europarl (4,395) + EuropeanaNewspaper (8,879) + \tueba (83,832)}
	
	\subsection{Word Embeddings}
	German is a highly inflected language compared to English or Chinese whose syntax is more analytic. For languages like German, the embedding of a single word (e.g.\ \textit{klein}) is dispersed across its various morphological and spelling variants (stem: \textit{klein} $\rightarrow$ \textit{kleiner}, \textit{kleinste}, \textit{kleine}, \textit{kleines}, \textit{kleinen}, \textit{kleinem}, \textit{Klein} etc.), therefore reducing the number of its samples and weakening its information value if not being lemmatized appropriately. On the other hand, languages with a rather analytical syntax show such  morphological variants to a lesser extent, if at all.
	We assume that this difference is the reason why their embeddings are of higher quality and therefore their performance in downstream tasks is many times higher than in less analytical languages.
	\begin{table}[htbp]
		\caption{Text Corpora}
		\begin{center}
			\begin{tabular}{|c|c|}
				\hline
				\textbf{Corpus} & \textbf{\textit{Sentences}} \\
				\hline\hline
				Leipzig40-2018&\textcolor{white}{0}40.00 Mio.\\
				WMT-2010-German&\textcolor{white}{0}19.36 Mio.\\
				\hline
				COW-2016&617.28 Mio.\\
				\hline
			\end{tabular}
			\label{tab-datasets-w2v}
		\end{center}
	\end{table}
	In order to mitigate this factor for the German language in its negative effect, we are therefore forced to use embeddings of higher quality. In the experimental setup of single training, we tackle this by using more text data. Table \ref{tab-datasets-w2v} lists the corpora we use for training our word embeddings. Leipzig40-2018 contains the largest possible extract from the so-called Leipzig Corpora Collection in 2018, which was generated by its maintainers on demand for our study, omitting any possible duplicate sentences. To increase the corpus size we combine this extract with WMT-2010-German forming our so-called \textit{LeipzigMT} corpus. Besides, we consider the COW-2016 corpus, arguably the largest text collection for German. This corpus contains not only a textbook-like language, as found for example in Wikipedia. Therefore, we assume that it fits well with the NER datasets used here, which in turn come from various sources (news, web, wikis, etc.). Both corpora are already preprocessed and split into sentences, containing words, numbers and punctuations. We do not remove punctuation marks, but separate them from words and numbers by surrounding them with spaces to avoid the introduction of variations with punctuation marks. In addition, as a preprocessing step, we write all words in lowercase to account for spelling and morphological variations.
	
	In a third variant of our experiment we deepen the optimization of resources by taking into account lemmatization and POS tagging in connection with writing words in lower case. While lemmatization increases the observation frequency of words, POS tagging allows a more correct specification of their syntactic roles in sentences and consequently differentiates individual observations that are included in the calculation of embeddings. On the other hand, lower case writing of words removes ambiguities, as they are induced in German especially by capitalization at the beginning of sentences.
	Table \ref{tab-w2v-params} shows the variations we use for this setup.
	
	We apply lemmatization and POS tagging in combination with writing words in lowercase to all resources before they are used in training. 
	These conversions are coupled with an exact conversion of the NER data sets in the respective experiment to avoid mismatches and to increase the overlap with the trained embeddings. Again, we use spaCy for these tasks and use its language models for lemmatization and POS tagging. Listing \ref{lst:lemma-pos} shows an example of this approach. 
	\begin{lstlisting}[frame=single, basicstyle=\scriptsize, caption=Example for Lemma \& POS, label=lst:lemma-pos]
raw sentence  : Kleine Kinder sind mutiger.
lemma         : Klein Kind sein mutig .
lemmapos      : Klein_ADJA Kind_NN sein_VAFIN mutig_ADJD ._$
lemmapos_lower: klein_ADJA kind_NN sein_VVFIN mutig_ADJD ._$
	\end{lstlisting}
	These conversions are intended to standardize any text input and thus to solve the above-mentioned problems in connection with morphological variations.
		
	\begin{table}[htbp]
		\caption{Embedding Variants per Experimental Setups}
		\begin{center}
			\begin{tabular}{|c||c|c|}
				\hline
				\textbf{Experimental Setup} & \textbf{\textit{Variant}}& \textbf{\textit{Features}} \\
				\hline\hline
				\textbf{Single Training}&1&lower\\
				\textbf{Joint Training}&1&lower\\
				\hline
				&2&lemma\\
				\textbf{Optimized}& 3& lemma\_lower\\
				\textbf{Training}& 4& lemmapos\\
				&5& lemmapos\_lower\\
				\hline	
			\end{tabular}
			\label{tab-w2v-params}
		\end{center}
	\end{table}

	\subsection{Training Parameters}
	To remain comparable with the baseline models on CoNLL \cite{glample2016} and GermEval \cite{nreimers2014}, we train the word embeddings with dimension 100\footnote{Lample et al.\ \cite{glample2016} use dimension 100 for English, but 64 for German. We increase this dimension to close the gap.}, window size of 8 and minimum word count threshold of 4, consequently, setting the LSTM dimension to 100 as well\footnote{For word2vec, we performed an extensive search on numerous embeddings with dimension values $ (50,100,150,200,300) $ along with minimum word count threshold and window size values in the range of $ [4,200] $ and $ [5,10] $, respectively. However, no major differences were observed in the final results.}. We choose dimension 25 for character-based embeddings and the final CRF-layer, and train the network in 100 epochs with a batch-size of 1 and dropout rate of 0.5. As an optimization method, we use the stochastic gradient descent with a learning rate of 0.005.
	Apart from fitting the LSTM dimension to 300 while using the 300-dimensional pretrained German fastText embeddings \cite{bojanowski2017enriching}, the model is fixed throughout our experiments to these settings. Any further sophisticated hyperparameter tuning (e.g.\ \textit{Population Based Training}) is left for future work.
	\section{Results}
	In this section, we present the results we obtained for our three experimental settings. 
	As described in \cite{Reimers2017ReportingSD}, we perform every experiment up to 6 times, starting from different random seeds, in order to arrive at  significant final values on the respective test dataset. We evaluate the NER results by using the official evaluation script from the shared task of CoNLL 2003. All our experiments were run on Nvidia's \textit{GTX 1080 Ti} GPUs.
	
	\subsection{Single Training}
	We compare our results with the current top performing models on CoNLL and GermEval. Table \ref{tab-single-training} shows the highest results we achieve on the single training setup (first experimental setting).
		
	\begin{table}[htbp]
		\caption{Single Training}
		\begin{center}
			\begin{tabular}{|c|c|c|l|}
				\hline
				\textbf{Data} & \textbf{Embeddings}& \textbf{Features}& \textbf{\textit{F-score} [\%]} \\
				\hline \hline
				CoNLL&\textit{pre-trained \textbf{Leipzig}}&wang2v&78.76 \cite{glample2016}\\
				\hline
				GermEval&\textit{pre-trained \textbf{UKP2014}}&word2v&75.9\textcolor{white}{0} \cite{nreimers2014}\\ 
				\hline\hline
				CoNLL&\textit{self-trained \textbf{LeipzigMT}}&wang2v&80.81\\
				CoNLL&\textit{self-trained \textbf{COW}}&wang2v&\textbf{83.29}\\
				\hline
				GermEval&\textit{self-trained \textbf{LeipzigMT}}&wang2v&81.97\\
				GermEval&\textit{self-trained \textbf{COW}}&wang2v&\textbf{83.14}\\
				\hline
				\tueba&\textit{self-trained \textbf{LeipzigMT}}&wang2v&88.95\\
				\tueba&\textit{self-trained \textbf{COW}}&wang2v&\textbf{89.26}\\
				\hline
			\end{tabular}
			\label{tab-single-training}
		\end{center}
	\end{table}
	
	We achieve an improvement throughout the datasets, outperforming all previous results on German neural NER, and establishing a new state-of-the-art on each of them. Increasing the corpus size by means of the LeipzigMT corpus displays a side-by-side performance increase on the CoNLL baseline. Increasing the corpus size further through the COW corpus gives us finally the best results on CoNLL. From this perspective, looking at the three data points for CoNLL (or GermEval), we observe a logarithmic growth of F-score as a function of the size of the underlying embedding corpus. Even larger corpora than the COW corpus are needed to further support this observation.
	
	On the side of training data, we observe a similar but more powerful behavior. On LeipzigMT, the increase of training data size from CoNLL to GermEval, and then to \tueba leads to an improvement of +1.16\% and +6.98\% in F-score. For COW this behavior re-emerges for \tueba, closing the gap to high-resource languages like English, and almost crossing the 90\% barrier on \tueba. Besides, we see that the larger train dataset \tueba does not heavily depend on the corpus size implying that it is beneficial to invest in annotation efforts.
	
	We also find that wang2vec generally performs better than word2vec. This shows that a task-specific embedding algorithm is important (in our case taking into account the syntax for NER).
	
	Last but not least, our experiments show that keeping information about capitalization can even downgrade the quality of word embeddings. Likewise, we  observe that integrating capitalization information as an additional input feature to our neural network does not lead to better results. We assume that this is due to the inflectional morphology of German, according to which nouns are capitalized at the beginning, in contrast to English, where mainly proper names (named entities) are written in this way.
	
	\subsection{Joint Training}
	As a first step towards joint training, we report the best results for fastText embeddings and compare them to UKP2014 embeddings, only using the two datasets from the baseline models. Next, we approach the full joint setup and perform the training on all German NER datasets. Starting from the results of the last section, we consider only COW for this setup. Table \ref{tab-joint-training} shows the top results for this setup.
	
	For fastText, we get the best results among all settings we examined (the results on single training were worse than for this setup). However, they are still below the ones with UKP2014, which themselves were trained with the original word2vec model back in 2014. This shows, that the fastText algorithm, being a promising extension of word2vec, does not suit well to our NER task, even though using a more informative vector space with 300 dimensions. Hence, we discard it for further experiments.
	
	For COW, the transfer learning on a single task works well and the performance for CoNLL and GermEval are improved further, lying slightly above the single training values. It can be noted that the final performance is more directed towards the low performing values. We assume that it depends more on the datasets with the lower single training performance (who make with $\sim37$\% a large part of the joint training dataset), as due to the data merging additional variety is introduced to the final training dataset. This makes the tasks more difficult and brings it closer to a real-world scenario. Still, the slightly improved performance indicates that the neural network is generalizing, and successfully performing \textit{task-related transfer learning on datasets}, i.e. the model is improving the same task on a heterogeneous dataset, given that it performs well on a single large homogeneous dataset.
	
	Overall, the results are promising; they indicate that we have a good candidate for applying a jointly trained tagger to large resources where the availability of labeled data is scarce.
	\begin{table}[htbp]
		\caption{Joint Training}
		\begin{center}
			\begin{tabular}{|c|c|c|l|}
				\hline
				\textbf{Data} & \textbf{Embeddings}& \textbf{Features}& \textbf{\textit{F-score} [\%]} \\
				\hline \hline
				CoNLL+GermEval&\textit{pre-trained \textbf{UKP2014}}&word2v&78.06\\
				CoNLL+GermEval&\textit{pre-trained \textbf{fastText}}&300dim&77.00\\
				\hline
				\hline
				\textbf{all}&\textit{self-trained \textbf{COW}}&wang2v&\textbf{83.47}\\
				\hline
			\end{tabular}
			\label{tab-joint-training}
		\end{center}
	\end{table}
	
	\subsection{Resource Optimization via Lemmatization \& POS tagging}
	In this final setup of resource optimization, we examine various constellations. Table \ref{tab-lemma-pos-result} reports the corresponding list of results.
	
	\begin{table}[htbp]
		\caption{Optimized Training via Lemma \& POS}
		\begin{center}
			\begin{tabular}{|c|c|c|l|}
				\hline
				\textbf{Data} & \textbf{Embeddings}& \textbf{Features}&\textbf{\textit{F-score} [\%]} \\
				\hline \hline
				&\textit{\textbf{LeipzigMT}}&lemma&82.57\\
				&\textit{\textbf{LeipzigMT}}&lemma\_lower&82.94\\
				&\textit{\textbf{LeipzigMT}}&lemmapos&81.22\\
				CoNLL&\textit{\textbf{LeipzigMT}}&lemmapos\_lower&81.20\\
				&\textit{\textbf{COW}}&lemma&\textbf{83.64}\\ 
				&\textit{\textbf{COW}}&lemma\_lower&83.14\\
				&\textit{\textbf{COW}}&lemmapos&82.38\\ 
				&\textit{\textbf{COW}}&lemmapos\_lower!&82.47\\ 
				\hline
				&\textit{\textbf{LeipzigMT}}&lemma&82.53\\
				&\textit{\textbf{LeipzigMT}}&lemma\_lower&82.47\\ 
				&\textit{\textbf{LeipzigMT}}&lemmapos&81.46\\
				GermEval&\textit{\textbf{LeipzigMT}}&lemmapos\_lower&81.05\\ 
				&\textit{\textbf{COW}}&lemma&\textbf{82.87}\\ 
				&\textit{\textbf{COW}}&lemma\_lower&82.53\\ 
				&\textit{\textbf{COW}}&lemmapos&81.96\\ 
				&\textit{\textbf{COW}}&lemmapos\_lower!&81.38\\		
				\hline
				&\textit{\textbf{LeipzigMT}}&lemma&88.50\\ 
				&\textit{\textbf{LeipzigMT}}&lemma\_lower&88.27\\ 
				&\textit{\textbf{LeipzigMT}}&lemmapos&87.85\\
				\tueba&\textit{\textbf{LeipzigMT}}&lemmapos\_lower!&87.83\\ 
				&\textit{\textbf{COW}}&lemma&89.08\\ 
				&\textit{\textbf{COW}}&lemma\_lower&\textbf{89.24}\\ 
				&\textit{\textbf{COW}}&lemmapos&88.43\\ 
				&\textit{\textbf{COW}}&lemmapos\_lower&88.02\\ 
				\hline
			\end{tabular}
			\label{tab-lemma-pos-result}
		\end{center}
	\end{table}
	
	Intuitively, using POS tagged sentences for training word embeddings may appear to be unusual, however, the results show a different picture. We get results very close to the top performances of the previous sections. A common pattern across all experiments can be detected. The variation of lemmatization on COW constantly delivers top scores for the three major datasets, and even produces the highest value for CoNLL across all setups. 
	Lemmatization performs comparatively better than lemmatization combined with POS tagging. This shows that dispersing the semantics of a given word across various roles it can take does not improve the quality of the final embeddings. Rather it is better to decrease the (redundant) varieties in the vector space by assembling in advance all morphological variants to a common base form, which only then is mapped to a common semantic vector.
	After lemmatization is performed, we can see that lower casing does not lead to a notable improvement. We assume that lemmatization already performs a good filtering of the raw text, making lower casing almost ineffective.
	
	Regarding the size of the corpus used for generating the word embeddings, we come to the conclusion, that lemmatization and POS tagging reduce the performance differences from previous sections which depended so far on the latter size. This confirms our assumption that the word2vec algorithm in its original form does not suit well to morphological rich languages. The results of this setup show that the values for LeipzigMT and COW now lie closer to each other, making the performance to some extent independent from the size of the embedding corpus. This is an important finding, giving rise to promising opportunities and applications for low-resource languages.

	\section{Conclusion \& Future Work} 
	In this paper, we performed a far reaching study on neural NER by example of a low-resource language like German. The study focused on a monolingual experimental setup. Nevertheless, the improved results pave the way for related languages with similar characteristics as German.
	
	There are various ways to improve existing neural models. Instead of just designing deeper and wider hybrid models, we showed the high importance of gathering and merging resources and how their careful optimization can eliminate the lack of progress. 
	In particular, we found out that increasing the size and improving the quality of raw corpora for word embeddings by applying morphological processing like lemmatization \& POS tagging leads to meaningful improvements.
	In addition, we demonstrated the effect of transfer learning by merging data sets for a joint training setup, which also produced good results and makes this approach a promising candidate for NER applications in the area of scarce resources of annotated data sets.
	
	Overall, we conducted the first comprehensive research for the German NER on all existing training data sets and resources, including the study of common pre-trained embeddings such as fastText. 
	In this context, we established a new state-of-the-art using all open source data sets for the German NER, which exceeds the 80\% F-score limit for the German NER and closes the gap to other high-resource languages such as English.
	
	For future work we plan to further refine the training process of word embedding and in particular to investigate how the performance of downstream tasks can become more independent of the size of embedding corpora using linguistic methods such as lemmatization and POS tagging. 
	To this end, we intent to examine the recently published ELMo embeddings \cite{Peters:2018} for German. 
	Finally, we will examine the role of the multilingual COW corpus for word embedding by example of other languages such as Dutch, French, Spanish and English.
	
	\section*{Acknowledgment}
	This work was funded by the German Research Foundations (DFG) as part of the \textit{BIOfid} project (DFG-326061700). We plan to upload our source code and the trained embeddings on GitHub for the research community.
	Special thanks goes to G.\ Lample for his directions on the procedure for training the embeddings, and to Prof.\ G.\ Heyer and F.\ Helfer for providing the extract of Leipzig40-2018 corpus.
	
	\bibliographystyle{IEEEtran}
	\bibliography{icmla2018_ahmed_mehler_arXiv}
	
\end{document}